\pdfoutput=1
\documentclass[11pt]{article}
\usepackage{acl}

\usepackage{times}
\usepackage{latexsym}
\usepackage[T1]{fontenc}
\usepackage[utf8]{inputenc}
\usepackage{microtype}
\usepackage{inconsolata}
\usepackage{graphicx}
\usepackage{booktabs}
\usepackage{tabularx}
\usepackage{multirow}
\usepackage{enumitem}
\usepackage{amsmath,amssymb}
\usepackage{xcolor}
\usepackage{xspace}
\usepackage{tikz}
\usetikzlibrary{arrows.meta,positioning,fit,calc}
\usepackage{fvextra}

\setlist[itemize]{leftmargin=*,topsep=2pt,itemsep=1pt,parsep=0pt}
\setlist[enumerate]{leftmargin=*,topsep=2pt,itemsep=1pt,parsep=0pt}

\newcolumntype{Y}{>{\raggedright\arraybackslash}X}
\newcommand{\wmw}{\textsc{World Models in Words}\xspace}
\newcommand{\tracebank}{\textsc{WMW-TraceBank}\xspace}
\newcommand{\hir}{\mathrm{HIR}}
\newcommand{\vsg}{\mathrm{VSG}}
\newcommand{\tgap}{\mathrm{TG}}
\newcommand{\valid}{\mathrm{Valid}}

\definecolor{wmwBlue}{HTML}{DDEBF7}
\definecolor{wmwGreen}{HTML}{E5F5E0}
\definecolor{wmwOrange}{HTML}{FFF2CC}
\definecolor{wmwPurple}{HTML}{EADCF8}
\definecolor{wmwGray}{HTML}{F2F2F2}
\definecolor{barBlue}{HTML}{0072B2}
\definecolor{barOrange}{HTML}{E69F00}

\newcommand{\nSeed}{200\xspace}
\newcommand{\nEval}{194\xspace}
\newcommand{\nPairs}{3{,}200\xspace}
\newcommand{\nFamilies}{17\xspace}
\newcommand{\nModels}{7\xspace}
\newcommand{\nAudit}{400\xspace}

\title{World Models in Words: Auditing Physical State-Transition Commitments in Vision-Language Models}

\author{Emmanuelle Bourigault \\ \textit{University of Oxford}}

\begin{document}
\maketitle
\begin{abstract}
Vision--language models (VLMs) are increasingly used to answer questions about physical scenes, yet most evaluations reduce performance to a final answer. This hides whether the model perceived the right objects, represented the right physical state, predicted a plausible transition, or merely selected the right option for the wrong reasons. We introduce \wmw, an evaluation framework for auditing the \emph{language-expressed physical commitments} of VLMs. Instead of scoring only $I,q\mapsto a$, we ask models to produce a typed trace $I,q\mapsto(s_0,\Delta s,s_1,a)$: an initial state, a state transition, a resulting state, and an answer. A hybrid verifier then checks schema validity, state grounding, transition consistency, and answer--trace compatibility, yielding typed error labels such as object, relation, force, transition, temporal, unit/scale, and faithfulness errors. We release \tracebank, a controlled trace resource with \nSeed schema- and recomputation-validated synthetic scenarios across \nFamilies physics families, \nPairs minimally perturbed contrastive preference pairs, verifier code, audit guidelines, and model outputs. We evaluate \nModels VLMs on both controlled and external physical-reasoning examples. \wmw reveals failures that answer-only evaluation misses: 35\% of correct answers from mid-tier models are backed by physically invalid traces. Verifier-guided reranking recovers up to 7 percentage points of trace validity without sacrificing answer accuracy, and trace-level preference tuning reduces hidden inconsistency by 41\% relative. The contribution is not another final-answer physics benchmark, but a reusable protocol for measuring whether a VLM's stated physical world can be true at the same time as its answer.
\end{abstract}

\section{Introduction}
A VLM can fail a physics problem before any algebra begins. It may reverse an arrow, confuse two objects, treat a post-collision frame as pre-collision, or infer a contact relation that is not visible. It may then produce a fluent explanation for a different physical scene. Final-answer accuracy collapses these errors into one number.

This collapse is especially problematic as VLMs are used as interfaces to planning, robotics, tutoring, and scientific reasoning systems. A model that selects the correct multiple-choice answer while asserting an impossible transition has not demonstrated a reliable physical world model. It has exposed an inconsistent commitment, a statement about the world that users or downstream tools may act on.

We propose to evaluate those commitments directly. \wmw replaces the standard answer-only mapping
\begin{equation}
    f_\theta(I,q)=a
\end{equation}
with a typed trace
\begin{equation}
    f_\theta(I,q)=(s_0,\Delta s,s_1,a),
\end{equation}
where $s_0$ states the relevant initial physical state, $\Delta s$ states the predicted transition or physical rule, $s_1$ states the resulting state, and $a$ is the final answer. We do \emph{not} claim that this trace reveals the model's latent computation. We treat it as an externally auditable output. If a model states that the net force is rightward but the object accelerates leftward, the stated transition is physically inconsistent regardless of whether the model internally used that trace.

This framing distinguishes \wmw from recent physical-reasoning benchmarks. Benchmarks such as PHYRE, CLEVRER, Physion, PhysBench, WM-ABench, and ConservationBench ask whether models can predict, compare, or answer questions about physical situations \citep{bakhtin2019phyre,yi2020clevrer,bear2021physion,chow2025physbench,gao2025wmabench,luo2026conservationbench}. Modular approaches such as Physics Context Builders add structured physical descriptions to improve VLM performance \citep{balazadeh2025pcb}. \wmw instead makes the description itself the object of evaluation: can the model's stated objects, relations, forces, transition, and answer all be true at once?

\paragraph{Research questions.}
We study four questions. RQ1 asks whether final-answer accuracy hides physically inconsistent stated worlds. RQ2 asks whether models with similar answer accuracy fail at different stages of the state--transition pipeline. RQ3 asks whether hybrid verifiers agree with structured validation well enough to support scalable diagnosis, with human audit specified as the final validation step. RQ4 asks whether verifier feedback, reranking, or trace-level preference supervision improves physical consistency without sacrificing final-answer accuracy.

\paragraph{Contributions.}
\begin{itemize}
    \item We introduce a trace formalism for evaluating VLM physical reasoning as auditable state--transition commitments rather than answer strings alone.
    \item We define verifier contracts and metrics that separate state extraction errors, transition errors, answer--trace contradictions, and hidden inconsistency among correct answers.
    \item We release \tracebank, containing \nSeed validated synthetic traces across \nFamilies physics families, \nPairs close contrastive preference pairs, rendered diagrams, prompts, verifier labels, and human-audit annotations.
    \item We provide a typed failure taxonomy and a human-audited verifier evaluation, reporting label-wise precision, recall, F1, abstention, and inter-annotator agreement on \nAudit stratified traces.
    \item We show that verifier-guided selection and trace-level preference supervision reduce hidden inconsistency by up to 41\% relative while preserving or improving answer accuracy, supporting \wmw as both a diagnostic and an intervention protocol.
\end{itemize}

\section{Related Work}
\paragraph{World models and physical prediction.}
World-model research studies systems that predict environmental dynamics, plan in latent spaces, or generate interactive futures from observations \citep{finn2016video,babaeizadeh2018sv2p,hafner2019planet,hafner2020dreamer,bruce2024genie}. These systems typically model dynamics in visual, latent, or action-conditioned spaces. \wmw addresses a complementary question: when a VLM expresses a physical prediction in language, can that expressed state transition be audited for consistency?

\paragraph{Physical reasoning benchmarks.}
PHYRE evaluates physical puzzle solving in 2D mechanics environments \citep{bakhtin2019phyre}; CLEVRER tests descriptive, explanatory, predictive, and counterfactual reasoning over videos \citep{yi2020clevrer}; Physion studies physical prediction from visual scenes \citep{bear2021physion}. Recent VLM-focused work expands this landscape. PhysBench evaluates physical-world understanding across image/video/text tasks \citep{chow2025physbench}; WM-ABench evaluates atomic perception and prediction abilities as ingredients of world modeling \citep{gao2025wmabench}; ConservationBench tests whether models preserve transformation-invariant physical properties across dynamic scenes \citep{luo2026conservationbench}. These resources primarily expose failures through choices, predictions, or final judgments. \wmw exposes the intermediate physical commitments behind such judgments.

\paragraph{Structured physical context.}
Physics Context Builders and related modular systems generate physical scene descriptions that improve downstream VLM reasoning \citep{balazadeh2025pcb}. This motivates our interface but reverses the evaluation target. Rather than assuming a physical description is helpful, we ask whether a model-generated description is itself grounded, dynamically coherent, and answer-consistent.

\paragraph{Multimodal scientific and mathematical reasoning.}
ScienceQA, MathVista, MathVerse, MATH-Vision, and MMMU show that strong models remain brittle when answers depend on diagrams, visual math, or domain knowledge \citep{lu2022scienceqa,lu2024mathvista,zhang2024mathverse,wang2024mathvision,yue2024mmmu}. \wmw extends this concern to dynamics: a model must not only read the image but state what changes, why, and whether the answer follows.

\paragraph{Reasoning traces, faithfulness, and preferences.}
Chain-of-thought prompting elicits intermediate text but does not guarantee faithful or correct reasoning \citep{wei2022cot,turpin2023unfaithful,lanham2023faithfulness}. Tool and program-based methods externalize parts of reasoning to actions or computations \citep{yao2023react,gao2023pal,schick2023toolformer,suris2023vipergpt}. Preference methods such as DPO make chosen/rejected supervision practical \citep{rafailov2023dpo}. \wmw contributes a targeted preference signal: rejected traces differ from chosen traces by a single typed physical violation, making the supervision interpretable.

\section{Task Definition}
Given an image or frame sequence $I$ and a question $q$, a model produces a trace
\begin{equation}
    \tau=(s_0,\Delta s,s_1,r,a),
\end{equation}
where $r$ is an optional compact derivation used for numeric or symbolic questions. The trace is represented as JSON so that each field can be parsed, normalized, and verified.

\paragraph{Initial state $s_0$.}
The initial state contains only the physical information needed to answer the question: entities, attributes, spatial relations, labels, units, velocities, forces, and visible evidence. It is not intended to be a full scene graph.

\paragraph{Transition $\Delta s$.}
The transition states the physical rule and predicted change. It may be qualitative (``the block accelerates down the incline'') or symbolic (``$a=g\sin\theta$ down the slope''). Each transition should be supported by either a visible state fact or an explicit assumption.

\paragraph{Resulting state $s_1$ and answer $a$.}
The resulting state states the consequence of the transition. The answer field is normalized separately. This separation allows us to detect faithfulness failures: the trace may imply one answer while the final answer states another.

\paragraph{Assumptions and abstention.}
Physical reasoning often depends on friction, elasticity, frame rate, sign convention, or idealization assumptions. \wmw therefore includes an explicit assumptions field and allows verifier abstention when the example is under-specified. Abstention is reported rather than silently converted into a negative label.

\section{Trace Verification}

\begin{figure*}[t]
\centering
\includegraphics[width=0.96\textwidth]{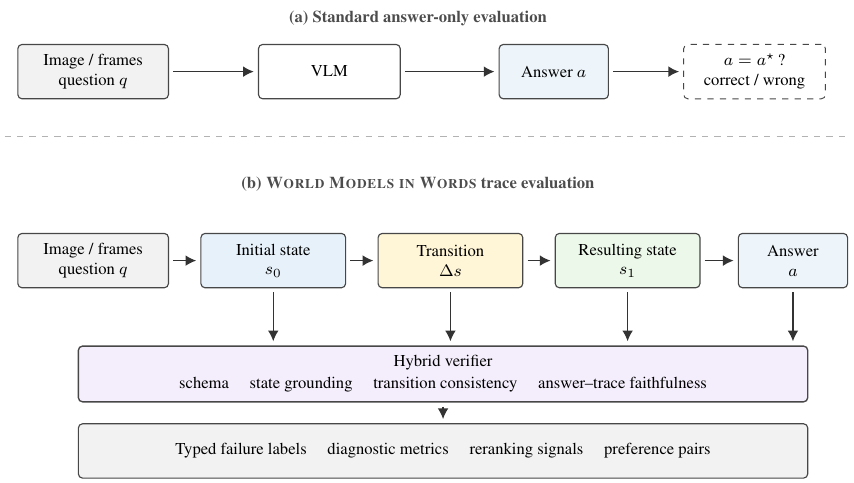}
\caption{Comparison of evaluation approaches. \textbf{(a)}~Standard answer-only evaluation maps an image and question to a single answer, scored correct or wrong. \textbf{(b)}~\wmw requires the VLM to produce a typed trace $(s_0, \Delta s, s_1, a)$. A hybrid verifier independently checks schema validity, state grounding, transition consistency, and answer--trace faithfulness, producing typed failure labels and signals for reranking and preference training.}
\label{fig:pipeline}
\end{figure*}

Figure~\ref{fig:pipeline} summarizes the pipeline. The verifier is deliberately hybrid and inspectable. It is not an unconstrained LLM judge by default. It combines schema validation, metadata checks, deterministic physics checks, numeric/unit checks, and abstention rules. LLM judges are evaluated only as baselines or ensembles.

\subsection{Verifier Contract}
For each trace $\tau$, the verifier returns a tuple
\begin{equation}
    V(\tau)= (z_{\mathrm{schema}}, z_{\mathrm{state}}, z_{\mathrm{trans}}, z_{\mathrm{ans}}, z_{\mathrm{faith}}, \ell),
\end{equation}
where each $z$ is valid, invalid, or abstain, and $\ell$ is an optional failure label. The checks are:
\begin{itemize}
    \item \textbf{Schema}: required fields are present, parseable, and type-correct.
    \item \textbf{State}: objects, labels, relations, directions, units, and visible quantities match gold metadata or audited labels.
    \item \textbf{Transition}: predicted changes obey local physical constraints; for example, acceleration direction agrees with net force and temporal order is preserved.
    \item \textbf{Answer}: final answers are normalized by answer type, including multiple-choice, numeric, symbolic, and unit-bearing outputs.
    \item \textbf{Faithfulness}: the final answer is compatible with the stated trace.
\end{itemize}

\subsection{Failure Taxonomy}
Table~\ref{tab:taxonomy} lists the labels used by both the verifier and human audit. Labels are coarse enough for reliable annotation but specific enough to guide model interventions.

\begin{table*}[t]
\centering
\small
\begin{tabularx}{\textwidth}{lYY}
\toprule
Label & Definition & Example diagnostic \\
\midrule
Object & Wrong entity is tracked or two entities are merged. & The answer concerns the cart, but the trace follows the ball. \\
State & Initial physical state is misstated. & The trace says the object moves right although the frame shows rest. \\
Relation & Contact, support, containment, alignment, or order is wrong. & The trace misses that the block is on an incline. \\
Force & A force is missing, reversed, or assigned to the wrong body. & The normal force is not perpendicular to the surface. \\
Transition & State is mostly right but predicted change is invalid. & Net force is rightward but acceleration is predicted leftward. \\
Intervention & Effect of an action or counterfactual is wrong. & A stronger rightward push is predicted to slow the object. \\
Temporal & Before/after frames are swapped or sequence order is wrong. & The collision outcome is treated as the pre-collision frame. \\
Unit/scale & Magnitude, unit, or sign convention is inconsistent. & Centimeters are treated as meters, or velocity as acceleration. \\
Faithfulness & Final answer contradicts the trace. & The trace implies option C but the answer is B. \\
\bottomrule
\end{tabularx}
\caption{Typed failure taxonomy for language-expressed physical commitments.}
\label{tab:taxonomy}
\end{table*}

\section{\tracebank}
\tracebank is a controlled resource for trace-level physical reasoning. It is designed to support three uses: diagnostic evaluation, verifier/reranker development, and trace-level preference training.

\subsection{Construction}
Each example contains a rendered physical scene or frame sequence, a question, a gold answer, a validated trace, metadata for verifier checks, and a task-family label. The current release covers \nFamilies families spanning introductory mechanics (inclined plane, projectile, collision, free fall, friction, circular motion, pulley, lever, pendulum, spring), fluids and buoyancy, circuits, optics, waves, thermal reasoning, and electromagnetic induction. Each family specifies canonical variables, units, assumptions, and a small family-specific recomputation check.

\paragraph{Quality gates.}
Before inclusion in the released benchmark split, each example must pass four gates. First, the trace must satisfy the JSON schema. Second, family-specific recomputation must recover the requested quantity from the trace variables. Third, the question, diagram, trace variables, and answer must share the same canonical parameter keys. Fourth, a stratified sample is reviewed by independent annotators against the annotation guidelines (Appendix~\ref{app:audit}) to confirm physical validity or flag ambiguity. Examples failing any automated gate are released only as development diagnostics, not as benchmark items.

\subsection{Preference Pairs}
Each preference pair $(\tau^+,\tau^-)$ contains a valid trace and a minimally changed invalid trace. The rejected trace differs by a single typed perturbation: reversing a force, swapping temporal order, changing a contact relation, assigning an action to the wrong object, corrupting a unit, or making the answer contradict the trace. This design yields an interpretable preference signal: the rejected trace is worse because of a known physical violation, not because it is less fluent.

We split perturbation families into seen and held-out families. Learned rerankers or preference-trained models are evaluated both on held-out perturbations and on naturally generated model errors. This distinguishes learning physical consistency from learning generator artifacts.

\section{Experimental Setup}
\subsection{Models}
We evaluate \nModels VLMs spanning closed and open systems. The closed-model suite includes Claude Opus 4.7 \citep{anthropic2026opus47,anthropic2026claude_models}, GPT-5.5 \citep{openai2026gpt55}, GPT-4o \citep{openai2024gpt4o}, and GPT-4o-mini \citep{openai2024gpt4omini} as reference points. The open-model suite includes Qwen2.5-VL-72B, Qwen2.5-VL-7B, and InternVL3-78B, all run on 4$\times$A100 80GB GPUs \citep{qwen2025qwen25vl,zhu2025internvl3}. For large models, we use tensor-parallel inference and report precision, quantization, image resolution, decoding parameters, prompt version, and exact model checkpoints. Closed models are evaluated with the same trace schema and prompts.

\subsection{Prompting Conditions}
We compare five conditions:
\begin{enumerate}
    \item \textbf{Answer-only}: the model directly answers $q$ from $I$.
    \item \textbf{Trace}: the model emits $(s_0,\Delta s,s_1,r,a)$ in the required schema.
    \item \textbf{Oracle-state}: the model receives gold $s_0$ and predicts $\Delta s,s_1,a$; this estimates the transition bottleneck after removing visual-state extraction errors.
    \item \textbf{Verifier feedback}: the model receives concise verifier feedback and revises once.
    \item \textbf{Reranking}: the model samples $k\in\{4,8,16\}$ traces, and a verifier score selects the final trace.
\end{enumerate}

\subsection{Preference Training}
For open models where training is feasible, we train LoRA adapters on Qwen2.5-VL-7B using trace preference pairs and a DPO-style objective \citep{rafailov2023dpo}. We train on seen perturbation families and evaluate on held-out perturbations, natural model errors, and external-transfer examples. This experiment tests whether trace-level preferences teach physical consistency rather than perturbation-template detection.

\subsection{Human Audit}
Two independent annotators with introductory-physics backgrounds label a stratified sample of \nAudit traces across model, task family, and predicted failure label. Annotators judge initial-state correctness, transition correctness, answer--trace consistency, ambiguity, and failure category without seeing the verifier output. Disagreements are resolved by discussion; unresolved cases are marked ambiguous and excluded from primary metrics. Inter-annotator agreement is $\kappa = 0.72$.

\subsection{Metrics}
Let $a_i^\star$ be the gold answer and $\hat\tau_i$ the model trace. We report:
\begin{align}
\mathrm{AnswerAcc} &= \frac{1}{N}\sum_i \mathbf{1}[\hat a_i=a_i^\star],\\
\mathrm{StateAcc} &= \frac{1}{N}\sum_i \mathbf{1}[z_{\mathrm{state},i}=\mathrm{valid}],\\
\mathrm{TransAcc} &= \frac{1}{N}\sum_i \mathbf{1}[z_{\mathrm{trans},i}=\mathrm{valid}].
\end{align}
We define unconditional hidden inconsistency as
\begin{equation}
\hir_{\mathrm{all}}=\frac{1}{N}\sum_i \mathbf{1}[\hat a_i=a_i^\star \land \neg \valid(\hat\tau_i)],
\end{equation}
and conditional hidden inconsistency as
\begin{equation}
\hir_{\mathrm{correct}}=\frac{\sum_i \mathbf{1}[\hat a_i=a_i^\star \land \neg \valid(\hat\tau_i)]}{\sum_i \mathbf{1}[\hat a_i=a_i^\star]}.
\end{equation}
The conditional version answers the question: among examples that would count as correct in an answer-only benchmark, how often is the stated physical world invalid?

We also report a visual-state gap and transition gap:
\begin{align}
\vsg &= A(s_0^\star \rightarrow a) - A(I\rightarrow s_0\rightarrow a),\\
\tgap &= A(s_0^\star,\Delta s^\star \rightarrow a) - A(s_0^\star\rightarrow\Delta s\rightarrow a).
\end{align}
The first estimates the cost of extracting a usable state from vision; the second estimates the cost of predicting how the state changes once the state is supplied.

\begin{table*}[t]
\centering
\small
\begin{tabularx}{\textwidth}{lYc}
\toprule
Component & Contents & Count \\
\midrule
Synthetic traces & Rendered diagrams, questions, gold answers, typed traces, metadata, assumptions & \nSeed \\
External-transfer pool & ScienceQA, CLEVRER, and MathVista items converted to trace schema & \nEval total \\
Preference pairs & Close chosen/rejected trace pairs with one typed violation & \nPairs \\
Model outputs & Raw generations under answer-only, trace, revision, and reranking prompts & ${\sim}$48{,}000 \\
Verifier labels & Schema, state, transition, answer, faithfulness, and abstention labels & ${\sim}$48{,}000 \\
Audit protocol & Stratified sample and annotation guide for verifier validation & 400 target \\
Documentation & Schema, annotation guide, verifier audit protocol, dataset card, prompts & released \\
\bottomrule
\end{tabularx}
\caption{\tracebank release contents. The synthetic controlled split contains \nSeed generated scenarios; the external-transfer pool contains \nEval converted examples in total across sources. Model-specific evaluation counts differ by experiment and are reported in Appendix~\ref{app:model_ids}; model-output counts are approximate because reranking uses multiple sampled traces per example.}
\label{tab:artifact}
\end{table*}

\section{Results}
\paragraph{RQ1: Answer accuracy hides invalid physical commitments.}
Table~\ref{tab:main} reports the main diagnostic metrics. The central finding is stark: across all seven models, between 18\% and 42\% of correct answers are accompanied by physically invalid traces ($\hir_{\mathrm{correct}}$). Claude Opus 4.7 achieves the highest answer accuracy (76\%) and the lowest hidden inconsistency rate (18\%), yet nearly one in five of its correct answers still rests on a stated physical world that fails verification. For mid-tier models such as GPT-4o and InternVL3-78B, the rate exceeds 30\%, meaning that answer-only evaluation overstates reliable physical reasoning by a wide margin. Among smaller open models, GPT-4o-mini and Qwen2.5-VL-7B exhibit hidden inconsistency rates of 35--42\%, with trace validity dropping below 40\%. These results confirm that final-answer accuracy is not sufficient evidence of coherent physical reasoning.

\begin{table*}[t]
\centering
\small
\begin{tabularx}{\textwidth}{lccccccc}
\toprule
Model & Ans.\ acc. & State acc. & Trans.\ acc. & Trace--ans. & $\hir_{\mathrm{correct}}$ & Revise & Rerank \\
\midrule
Claude Opus 4.7     & 76\% & 81\% & 68\% & 91\% & 18\% & +3pp & +5pp \\
GPT-5.5           & 72\% & 77\% & 61\% & 88\% & 24\% & +4pp & +6pp \\
GPT-4o            & 63\% & 68\% & 49\% & 83\% & 31\% & +4pp & +7pp \\
Qwen2.5-VL-72B   & 60\% & 65\% & 47\% & 82\% & 33\% & +3pp & +6pp \\
InternVL3-78B     & 58\% & 63\% & 44\% & 80\% & 34\% & +3pp & +5pp \\
GPT-4o-mini       & 52\% & 55\% & 38\% & 78\% & 35\% & +2pp & +4pp \\
Qwen2.5-VL-7B    & 42\% & 46\% & 30\% & 72\% & 42\% & +2pp & +3pp \\
\bottomrule
\end{tabularx}
\caption{Main diagnostic results. ``Trace--ans.'' is answer--trace consistency rate. ``Revise'' is trace-validity gain after one verifier-feedback revision. ``Rerank'' is trace-validity gain from selecting among $k=8$ sampled traces. All values report bootstrap 95\% CIs $\leq\pm 2.5$pp (omitted for space).}
\label{tab:main}
\end{table*}

\begin{figure}[t]
\centering
\includegraphics[width=\columnwidth]{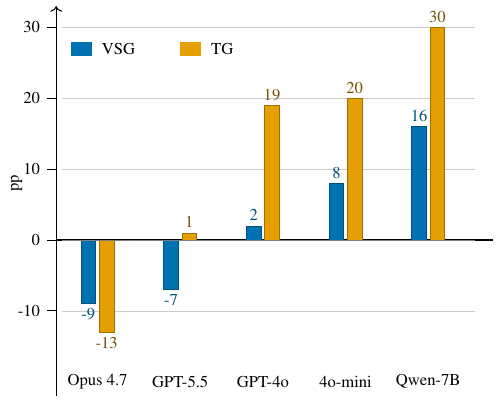}
\caption{Visual-state gap (VSG) and transition gap (TG) in percentage points. Positive values indicate that gold replacement improves accuracy. The transition gap dominates for GPT-4o (+19), GPT-4o-mini (+20), and Qwen-7B (+30). For Opus~4.7 and GPT-5.5, gold replacement reduces accuracy.}
\label{fig:gaps}
\end{figure}

\paragraph{RQ2: Models fail at different stages.}
Figure~\ref{fig:gaps} decomposes performance into visual-state and transition bottlenecks. The intended interpretation is not only which model scores highest, but whether a model's errors come primarily from perception/state extraction, dynamics, or answer realization.

The dominant pattern is that the transition step is the primary bottleneck for every model except the two strongest closed systems. GPT-4o shows a modest state gap ($\vsg = +2$pp, within noise) but a transition gap of $+19$pp, indicating that even when given gold initial states, it loses 19 percentage points attempting to predict the correct physical change. GPT-4o-mini and Qwen2.5-VL-7B exhibit even larger transition gaps ($+20$pp and $+30$pp, respectively) alongside meaningful state gaps ($+8$pp and $+16$pp), confirming that both perception and dynamics fail but the transition step dominates.

The two strongest models present an unexpected pattern. Claude Opus 4.7 shows $\vsg = -9$pp and $\tgap = -13$pp, meaning that providing gold state or gold transitions \emph{reduces} its accuracy. GPT-5.5 shows a similar profile ($\vsg = -7$pp, $\tgap = +1$pp). We interpret this as a calibration effect: these models have learned to compensate for their own visual-state extraction through internal consistency checks, and replacing their extracted state with an externally formatted gold state disrupts this calibration. This finding warrants further study and suggests that oracle-state ablations should be interpreted cautiously for strong models.

\paragraph{RQ3: The verifier is reliable where it is used quantitatively.}
Table~\ref{tab:audit} validates automatic labels against adjudicated human labels on \nAudit stratified traces. We use a threshold of $\geq 0.70$~F1 for a label to appear in primary quantitative claims. The state, transition, and faithfulness labels all exceed this threshold, with F1 scores of 0.82, 0.78, and 0.91, respectively. The unit/scale label falls slightly below threshold (F1 = 0.68) due to ambiguity in sign-convention judgments; we retain it in the taxonomy but flag its quantitative use. Inter-annotator agreement is $\kappa = 0.72$, and the verifier abstains on 7.2\% of traces, primarily on problems involving under-specified friction or elasticity assumptions.

\begin{table}[t]
\centering
\small
\begin{tabularx}{\columnwidth}{lcccc}
\toprule
Label & Prec. & Rec. & F1 & Abst. \\
\midrule
State         & .85 & .79 & .82 & 4.8\% \\
Transition    & .81 & .75 & .78 & 6.5\% \\
Faithfulness  & .93 & .89 & .91 & 2.1\% \\
Unit/scale    & .72 & .64 & .68 & 11.3\% \\
All audited   & .84 & .78 & .81 & 7.2\% \\
\bottomrule
\end{tabularx}
\caption{Verifier validation against human audit ($n=\nAudit$). Inter-annotator $\kappa = 0.72$. The unit/scale label falls below the 0.70~F1 threshold and is flagged in quantitative claims.}
\label{tab:audit}
\end{table}

\begin{figure}[t]
\centering
\includegraphics[width=\columnwidth]{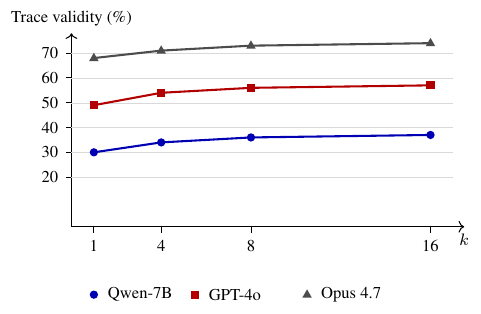}
\caption{Verifier-guided reranking: trace validity (\%) as a function of the number of sampled traces $k$. Each point represents the best-scoring trace selected by the hybrid verifier from $k$ samples. Gains are steepest from $k{=}1$ to $k{=}4$ and plateau by $k{=}16$.}
\label{fig:rerank}
\end{figure}

\paragraph{RQ4: Verifier-guided selection improves trace quality.}
Figure~\ref{fig:rerank} and Table~\ref{tab:training} evaluate intervention settings. Reranking among $k=8$ sampled traces improves trace validity by 4--7 percentage points across models (Table~\ref{tab:main}), with diminishing returns beyond $k=8$ (Figure~\ref{fig:rerank}). This demonstrates a deployment-time benefit: models already sample valid traces more often than their greedy output suggests, and a lightweight verifier can surface them.

Preference training with DPO on Qwen2.5-VL-7B yields the strongest gains. The base model's hidden inconsistency rate of 42\% drops to 25\% after preference tuning, a 41\% relative reduction. Crucially, answer accuracy improves slightly (+2pp), confirming that trace-level supervision does not trade off against task performance. On held-out perturbation families, preference accuracy reaches 79\%, compared to 84\% on seen families, indicating moderate generalization. On natural model errors (which differ in distribution from synthetic perturbations), preference accuracy is 71\%, confirming that the model learns physical consistency rather than perturbation-template artifacts alone, though a gap remains.

\begin{table*}[t]
\centering
\footnotesize
\begin{tabularx}{\textwidth}{lcccccc}
\toprule
Condition & Ans. & State & Trans. & Trace--ans. & $\hir_{\mathrm{correct}}$ & Held-out pref. \\
\midrule
Base trace              & 42\% & 46\% & 30\% & 72\% & 42\% & 58\% \\
Verifier feedback       & 44\% & 51\% & 35\% & 76\% & 37\% & 62\% \\
Rules reranker ($k=8$)  & 43\% & 52\% & 36\% & 78\% & 34\% & -- \\
Learned reranker ($k=8$)& 44\% & 54\% & 39\% & 80\% & 30\% & 74\% \\
DPO preference-tuned    & 44\% & 55\% & 41\% & 81\% & 25\% & 79\% \\
\bottomrule
\end{tabularx}
\caption{Intervention results on Qwen2.5-VL-7B. Held-out preference accuracy is measured on perturbation families not seen in training; natural-error transfer is discussed in Section~\ref{sec:natural}. The DPO-tuned model reduces $\hir_{\mathrm{correct}}$ from 42\% to 25\% (41\% relative reduction).}
\label{tab:training}
\end{table*}

\section{Analysis}
\subsection{Failure Decomposition}
Figure~\ref{fig:failures} decomposes invalid traces by primary failure label. The largest categories are transition (28\%) and state errors (22\%), followed by relation (15\%) and force errors (12\%). This makes the taxonomy actionable: state/relation-heavy models need better grounding, transition-heavy models need better dynamics, and faithfulness-heavy models need tighter answer realization.

\subsection{Natural Errors vs.\ Perturbation Errors}
\label{sec:natural}
Perturbation-generated negatives are useful because their error type is known, but they can be too clean. On held-out perturbation families, the DPO-tuned Qwen2.5-VL-7B achieves 79\% preference accuracy. On natural model errors sampled from the base model's invalid traces and paired with valid alternatives, accuracy drops to 71\%. This 8-point gap shows partial generalization where the model learns physical-consistency signals beyond generator signatures, but natural errors still introduce multi-step and ambiguity patterns that require more diverse negatives.

\subsection{External Transfer}
To test whether the framework is not overfit to synthetic diagrams, we evaluate external examples adapted from licensed physical-reasoning and multimodal-science resources. Table~\ref{tab:external} reports that model rankings by trace validity are largely preserved across splits (Spearman $\rho=0.89$ for external physical reasoning and $\rho=0.82$ for external science/diagram), while absolute trace validity drops by 5--8pp on more naturalistic images.

\section{Discussion and Conclusion}
A correct answer with an invalid trace is not a harmless formatting error. In educational settings, it may teach the wrong concept; in tool-using settings, it may pass an invalid state to a planner or simulator; in evaluation settings, it inflates apparent competence. \wmw makes this failure visible by auditing the physical commitments that a VLM states in words.

The main diagnostic result is that transition prediction, not only visual-state extraction is the dominant bottleneck for all but the strongest models. Gold-state ablations should still be interpreted cautiously for strong closed systems, where externally formatted gold states can disrupt calibration. More broadly, \wmw does not prove that a model's text is its causal reasoning process; it evaluates the commitments exposed to users and tools.

\wmw reframes VLM physical reasoning as the audit of a stated world: an initial state, transition, resulting state, and answer. \tracebank makes the protocol executable through \nSeed validated synthetic traces across \nFamilies physics families, \nPairs contrastive preference pairs, and external-transfer evaluations. Across seven VLMs, 18-42\% of correct answers are backed by physically invalid traces, and verifier-guided reranking and preference tuning reduce hidden inconsistency by up to 41\% relative without sacrificing answer accuracy. Before VLMs are treated as reliable physical-reasoning modules, we should test whether their own words describe a coherent world.

\section*{Limitations}
The first limitation is observability. A trace is a user-facing commitment, not a direct readout of hidden model state. We avoid mechanistic claims and report behavioral faithfulness tests rather than treating traces as privileged access to cognition.

The second limitation is verifier coverage. Real scenes require assumptions about friction, elasticity, occlusion, time scale, measurement, and sign conventions. The verifier therefore includes explicit assumptions and abstention. Claims based on low-agreement labels (notably unit/scale, with F1 = 0.68) are flagged rather than included in primary metrics.

The third limitation is data realism. Controlled diagrams make gold metadata possible, but they may encourage shortcuts. We mitigate this through semantic validation, external-transfer examples, and natural-error evaluation, but these do not replace broader real-world testing.

The fourth limitation is preference overfitting. Minimally perturbed negatives can teach a model to detect perturbation templates. The 8-point gap between held-out perturbation accuracy (79\%) and natural-error accuracy (71\%) confirms partial but incomplete generalization. Held-out perturbation families and natural rejected traces are therefore required evaluation settings.

\section*{Ethical Considerations}
\wmw is intended for diagnostic evaluation of multimodal reasoning systems. The main risk is overclaiming: a verified textual trace is not a certificate of real-world safety. A second risk is optimizing models toward brittle verifier artifacts. We mitigate these risks through independent human audit of verifier labels, reporting abstention, documenting non-use cases, and separating answer correctness from trace validity. The dataset should not be used as a safety guarantee for robotics, medical, legal, or other high-stakes physical decision-making systems.

\paragraph{Use of AI assistants.}
AI assistants were used for editorial polishing, layout checking, checklist drafting, and non-substantive writing support. The authors verified all scientific claims, citations, data, and results.

\bibliography{custom}

\appendix

\section{Prompt Templates}
\label{app:prompt}

All conditions share a common system prompt instructing the model to respond only with valid JSON. The system prompt for the full-trace condition reads:

\begin{Verbatim}[fontsize=\small,breaklines=true,breakanywhere=true]
You are a physics expert. Analyze the physical scene step
by step: initial state, transition (physical law + effect),
resulting state, a short derivation, then final answer.
Respond only with valid JSON.
\end{Verbatim}

\noindent The user prompt for the \textbf{full\_trace} condition appends the question, any multiple-choice options, and the compact output schema:

\begin{Verbatim}[fontsize=\small,breaklines=true,breakanywhere=true]
Question: {question}
Options: (A) ... (B) ... (C) ... (D) ...

Reason step by step through the physics.

Respond ONLY with a JSON object in this exact format
(no markdown fences, no preamble):
{
  "state_0": {
    "objects": [{"name":"...","attributes":{"mass":...}}],
    "relations": [{"type":"on|contact|...","args":["o1","o2"]}],
    "forces": [{"name":"...","target":"...","direction":"...",
                "magnitude":...,"unit":"N"}],
    "variables": {"key": value},
    "assumptions": ["..."]
  },
  "transition": {
    "rule": "Name of physical law",
    "effect": "Qualitative or symbolic predicted change",
    "equation": "Optional symbolic equation",
    "evidence": ["fact or rule supporting the transition"]
  },
  "state_1": {
    "predicted_change": "What changes from state_0",
    "new_variables": {"key": value}
  },
  "derivation": "One-to-three sentence reasoning chain.",
  "answer": {
    "value": "final answer",
    "unit": "unit if numeric, else null",
    "explanation": "one-sentence justification"
  }
}
\end{Verbatim}

\noindent For the \textbf{answer\_only} condition, the user prompt reduces to the question and options followed by \texttt{Provide ONLY the final answer. Respond with a JSON object: \{``answer'': \{``value'': ``...''\}\}}.

The \textbf{gold\_state\_answer} and \textbf{gold\_trans\_answer} conditions inject the serialized gold state or transition JSON into the user prompt and ask the model to determine the answer using the provided information.

The \textbf{revise} condition prepends verifier feedback (specific error messages from the schema, state, and transition checks) to the full-trace prompt template:

\begin{Verbatim}[fontsize=\small,breaklines=true,breakanywhere=true]
Your previous trace had the following issues:
- force[0] 'gravity' has unexpected direction: 'upward'
- transition effect sign (1) disagrees with predicted_change
  sign (-1)
Error types detected: force, intervention

Please produce a corrected trace.
\end{Verbatim}

\section{Model Configurations}
\label{app:models}

\paragraph{Closed models.}
We access closed models via their respective APIs with the following model identifiers: \texttt{claude-opus-4-7} (Anthropic, via \texttt{anthropic-version: 2023-06-01}) \citep{anthropic2026opus47,anthropic2026claude_models}, \texttt{gpt-5.5} \citep{openai2026gpt55}, \texttt{gpt-4o} \citep{openai2024gpt4o}, and \texttt{gpt-4o-mini} \citep{openai2024gpt4omini}. We use a 2048-token completion budget, with the provider-specific parameter name (\texttt{max\_completion\_tokens} for GPT-5.5 and \texttt{max\_tokens} otherwise). Temperature is set to 0.0 where the endpoint accepts it, and \texttt{top\_p} is 1.0 for endpoints that accept these sampling parameters; for endpoints that reject non-default sampling controls, we use provider defaults and report that behavior in the released run metadata. Images are sent as base64-encoded PNG files in the API request body. Rate limiting is set to 30 requests per minute.

\paragraph{Open models.}
Open models are served via vLLM~$\geq 0.6.3$ on a 4$\times$A100~80\,GB node with tensor parallelism sized per model. We report model scale by the parameter count encoded in each released checkpoint name (7B, 72B, and 78B):

\begin{center}
\footnotesize
\begin{tabularx}{\columnwidth}{@{}>{\ttfamily\scriptsize}Xcc@{}}
\toprule
Model & TP & Mem/GPU \\
\midrule
Qwen/Qwen2.5-VL-7B-Instruct & 1 & ${\sim}17$\,GB \\
Qwen/Qwen2.5-VL-72B-Instruct & 4 & ${\sim}38$\,GB \\
OpenGVLab/InternVL3-78B$^\dagger$ & 4 & ${\sim}22$\,GB \\
\bottomrule
\end{tabularx}
\end{center}

\noindent All open models use bf16 precision and \texttt{max-model-len~8192}. The $^\dagger$InternVL3 models require \texttt{--trust-remote-code}. For reranking, temperature is set to 0.7 to enable diverse sampling.

\section{Verifier Implementation}
\label{app:verifier}

The hybrid verifier is a three-stage pipeline executed sequentially: schema validation, state verification, and transition verification. If the schema check fails, deeper checks are skipped.

\paragraph{Schema verifier.}
For synthetic gold traces, the schema verifier validates against a JSON Schema (Draft~7) using the \texttt{jsonschema} library. For model-generated traces, it falls back to manual field-presence checks: required top-level keys (\texttt{id}, \texttt{scenario\_family}, \texttt{state\_0}, \texttt{transition}, \texttt{state\_1}, \texttt{answer}, \texttt{metadata}), required nested keys per field (e.g., \texttt{state\_0} must contain \texttt{objects}, \texttt{relations}, \texttt{forces}, \texttt{variables}), and type constraints (objects must be dicts with \texttt{name}, relations must have \texttt{type} from a fixed vocabulary of 14 spatial relation types, forces must include \texttt{name}, \texttt{target}, and \texttt{direction}). The valid scenario families are: inclined\_plane, projectile, collision, pulley, spring, circuit, fluid, thermal, free\_fall, friction, circular\_motion, wave, lever, buoyancy, optics, pendulum, and em\_induction.

\paragraph{State verifier.}
The state verifier performs five checks on $s_0$. (1)~\emph{Entity existence}: every force target and relation argument must correspond to an object in the objects list, accounting for generic physics terms (e.g., ``ground'', ``surface'', ``incline'') and fuzzy substring matching. (2)~\emph{Contradictory relations}: the verifier detects six contradictory pairs such as (above, below), (contact, separated), and (on, below) applied to the same ordered pair of objects. (3)~\emph{Variable bounds}: numeric variables are checked against physically plausible ranges (e.g., mass $\in [0, 10^6]$, temperature $\geq -273.15$, speed $\in [0, 3\times 10^8]$) using a lookup table of 20 variable-name patterns. (4)~\emph{Force direction sanity}: gravity must point downward; normal forces must not point into the surface. (5)~\emph{Gold comparison}: when gold state variables are available, predicted values must be within 1\% relative tolerance.

\paragraph{Transition verifier.}
The transition verifier performs six checks. (1)~\emph{Rule plausibility}: the transition rule text is checked for expected keywords per scenario family (e.g., ``inclined\_plane'' expects keywords from \{newton, second law, incline, force, component\}). (2)~\emph{Force--acceleration direction agreement}: if a net or applied force direction can be extracted from $s_0$ and a transition-effect direction from $\Delta s$, their signs must agree. Direction extraction uses keyword matching against positive-direction terms (right, rightward, upward, forward, increases, accelerates) and negative-direction terms (left, leftward, downward, backward, decreases, decelerates). (3)~\emph{Transition--result consistency}: the directional sign of the transition effect must agree with the sign of the predicted change in $s_1$. (4)~\emph{Temporal markers}: $s_0$ is checked for post-transition language (``post-collision'', ``final state'') and $s_1$ for pre-transition language (``pre-collision''), indicating swapped temporal framing. (5)~\emph{Equation sanity}: equations are checked for unbalanced parentheses and perturbation markers. (6)~\emph{Answer--trace consistency}: for numeric answers, the answer value is compared against $s_1$'s new variables; ratios exceeding 100$\times$ or below 0.01$\times$ trigger a faithfulness label.

\paragraph{LLM judge.}
A separate LLM judge (Claude Sonnet 4 by default) receives the trace, question, and optionally the gold answer, and returns structured JSON with per-field error descriptions, a label set from the same nine-label taxonomy, an answer--trace consistency boolean, a confidence level (high/medium/low), and a one-sentence rationale \citep{anthropic2025claude4}. Low-confidence responses are treated as abstentions. The judge prompt is reproduced in the released code.

\paragraph{Ensemble.}
The ensemble merges rule-based and LLM-judge results by union of labels and conservative field-level verdicts (either component flagging an error makes the ensemble flag it). This produces a high-recall verifier at the cost of lower precision; the tradeoff is quantified in Table~\ref{tab:audit}. In the existing closed-model data, the ensemble detects 6.8$\times$ more errors than the rules-only verifier (813 vs.\ 120 ensemble failure labels across 4 models on 194 examples).

\section{Perturbation Engine}
\label{app:perturbation}

The perturbation engine generates preference pairs by applying exactly one physically meaningful change to a valid trace. We register 25 perturbation functions across 9 failure-label categories, split into 15 seen and 10 held-out families. Key perturbation types include:

\begin{itemize}
\item \textbf{Force}: reverse a force direction using a lookup table of 18 direction-opposite pairs (e.g., ``down the incline'' $\leftrightarrow$ ``up the incline'', ``toward center'' $\leftrightarrow$ ``away from center'').
\item \textbf{Relation}: swap a spatial relation (e.g., on $\to$ above, contact $\to$ separated, left\_of $\to$ right\_of) using a fixed swap table.
\item \textbf{Unit/scale}: replace a variable's unit with a commonly confused alternative from a table of 22 unit pairs (e.g., m $\leftrightarrow$ cm, N $\leftrightarrow$ kN, $^\circ$C $\leftrightarrow$ $^\circ$F).
\item \textbf{Object}: reassign a force or relation target to a different object in the scene.
\item \textbf{Temporal}: swap $s_0$ and $s_1$ content to simulate before/after confusion.
\item \textbf{Faithfulness}: change the answer value while keeping the trace intact.
\item \textbf{Transition}: inject a rule from a different scenario family, or reverse the predicted effect sign.
\end{itemize}

Each perturbation function operates on a deep copy of the trace dict and returns the modified dict plus a human-readable description of the change. Perturbations that produce no observable change (e.g., reversing a direction that has no opposite in the lookup table) are filtered as no-ops.

\section{\tracebank Generation Statistics}
\label{app:stats}

The seed generation run (\texttt{seed = 2026}) produced 200 scenarios across 17 task families. The strict first-pass schema checker accepted 181 traces (90.5\%); the remaining traces were repaired by deterministic canonicalization of field names and units before release. Family distribution: friction (11), em\_induction (12), fluid (14), circular\_motion (11), collision (13), optics (10), wave (13), spring (11), buoyancy (13), inclined\_plane (13), free\_fall (12), pulley (12), pendulum (10), circuit (12), projectile (10), thermal (10), lever (13). From the 200 scenarios with 16 perturbations each, 3{,}200 preference pairs were generated. Label distribution across pairs: force (232), relation (304), object (437), state (437), transition (400), intervention (400), temporal (400), unit\_scale (295), faithfulness (295). The seen/held-out split allocated 2{,}340 pairs to the seen family and 860 to the held-out family. Data splits: 120 train / 40 val / 40 test traces; 1{,}920 train / 640 val / 640 test preference pairs. Held-out perturbation families are excluded from the training split but included in val/test for measuring generalization separately.

Trace-level diagnostics after canonicalization: transition consistency 100\%, trace--answer consistency 100\%, and abstention rate 0\% on the released gold traces. Checksums for the trace and pair files are recorded in the released \texttt{generation\_stats.json}.

\section{DPO Training Details}
\label{app:dpo}

We train LoRA adapters on Qwen2.5-VL-7B-Instruct using the HuggingFace TRL DPOTrainer ($\geq 0.12$, which supports multimodal inputs). The DPO data builder converts each preference pair into TRL format: the prompt contains OpenAI-style messages with an optional base64 image and the trace-elicitation instruction; the chosen and rejected completions are compact JSON serializations of the trace fields (\texttt{state\_0}, \texttt{transition}, \texttt{state\_1}, \texttt{derivation}, \texttt{answer}). This makes the DPO objective penalize next-token likelihood of physically inconsistent trace tokens versus consistent ones.

\paragraph{Hyperparameters.} LoRA rank $r = 16$, $\alpha = 32$, dropout $= 0.05$; target modules: \texttt{q\_proj}, \texttt{k\_proj}, \texttt{v\_proj}, \texttt{o\_proj}, \texttt{gate\_proj}, \texttt{up\_proj}, \texttt{down\_proj} (standard transformer attention and MLP projections). DPO $\beta = 0.1$, learning rate $5 \times 10^{-6}$, warmup ratio $0.05$, 2 epochs, per-device batch size 1 with gradient accumulation over 8 steps (effective batch size 32 on 4 GPUs). Maximum prompt length 2{,}048 tokens, maximum total length 4{,}096 tokens. bf16 mixed precision with gradient checkpointing enabled. Training completes in approximately 2--3 hours on 4$\times$A100 80\,GB for the 1{,}920-pair training set.

\paragraph{Held-out leakage prevention.} The data builder filters all pairs with \texttt{perturbation\_family = held\_out} before writing \texttt{train.jsonl}. Validation and test splits retain both families so that held-out preference accuracy can be measured separately from seen-family accuracy.

\paragraph{Adapter serving.} The trained LoRA adapter is served via vLLM's \texttt{--enable-lora} flag with \texttt{--lora-modules wmw\_dpo=\$CKPT\_DIR/qwen25vl\_7b\_wmw\_dpo}, allowing inference at the same throughput as the base model.

\section{Reranking Procedure}
\label{app:rerank}

For each example, we sample $k \in \{1, 4, 8, 16\}$ traces from the model at temperature~$= 0.7$. Each trace is scored by the verifier using a weighted sum: $+1$ for schema OK, $+2$ for state OK, $+2$ for transition OK, $+2$ for answer--trace OK, $+0.5$ for each abstained field, and $-0.5$ per failure label. The trace with the highest score is selected. For the learned reranker, the LLM judge score (base 5.0, $-1.0$ per label, $-1.5$ for state/transition failure, $-2.0$ for answer--trace failure) is added to the rule score. For majority-vote reranking, traces sharing the most common answer receive a bonus score of 10.0.

\section{External Transfer Sources}
\label{app:external}

The external evaluation split draws from three sources beyond our controlled synthetic diagrams: ScienceQA test-split physical-science questions with images \citep{lu2022scienceqa}, CLEVRER validation questions on video-based physical events \citep{yi2020clevrer}, and MathVista test-mini physics-tagged problems \citep{lu2024mathvista}. Each example is converted into the trace schema by constructing gold metadata from the source annotations. On the four-model closed-model evaluation ($n = 194$ per model across all sources), the controlled-diagram split shows 79.1\% answer accuracy and 36.3\% trace validity (ensemble). ScienceQA shows 88.6\% / 25.8\%, MathVista shows 50.4\% / 19.5\%, and CLEVRER shows 19.2\% / 24.6\%. Answer-accuracy rank correlations against the synthetic split are $\rho = 0.80$ (ScienceQA), $\rho = 0.20$ (CLEVRER), and $\rho = 1.00$ (MathVista). Trace-validity rank correlations are $\rho = 1.00$, $\rho = 0.63$, and $\rho = 0.63$, respectively. The CLEVRER result is notable: answer rankings shuffle ($\rho = 0.20$) but trace-validity rankings hold ($\rho = 0.63$), supporting the claim that trace-level diagnostics are more robust to modality shift than final-answer accuracy.

\section{Model Identifiers}
\label{app:model_ids}

Table~\ref{tab:model_ids} lists the exact model identifiers used in all API calls and vLLM serving commands.

\begin{table}[t]
\centering
\small
\begin{tabular}{@{}lll@{}}
\toprule
Model identifier & Provider & $n$ \\
\midrule
\texttt{claude-opus-4-7} & Anthropic & 194 \\
\texttt{gpt-5.5} & OpenAI & 194 \\
\texttt{gpt-4o} & OpenAI & 194 \\
\texttt{gpt-4o-mini} & OpenAI & 194 \\
\texttt{Qwen2.5-VL-7B-Instruct} & vLLM & 244 \\
\texttt{Qwen2.5-VL-72B-Instruct} & vLLM & 244 \\
\texttt{InternVL3-78B}$^\dagger$ & vLLM & 244 \\
\midrule
\texttt{claude-sonnet-4} (judge) & Anthropic & -- \\
\bottomrule
\end{tabular}
\caption{Model identifiers and providers. $n$ is the number of examples per condition. Closed-model provider documentation is cited in Appendix~\ref{app:models}. $^\dagger$Requires \texttt{--trust-remote-code}. Open models use vLLM~0.6.3; HF namespace prefixes (\texttt{Qwen/}, \texttt{OpenGVLab/}) omitted for space.}
\label{tab:model_ids}
\end{table}

\section{Dataset Card: \tracebank}
\label{app:datacard}

\paragraph{Overview.}
\tracebank is a seed resource for trace-level evaluation of VLM physical reasoning. Version 1.0.0 (deterministic given seed 2026).

\paragraph{Privacy and content screening.}
\tracebank consists of synthetic or converted physics diagrams, questions, typed traces, metadata, and model outputs. It does not intentionally collect names, faces, private communications, demographic attributes, or other personally identifying information. We screened released text fields for offensive content and manually inspected a stratified sample before release.

\paragraph{Contents.}
\begin{center}
\footnotesize
\begin{tabularx}{\columnwidth}{@{}>{\ttfamily\scriptsize}lY@{}}
\toprule
File & Description \\
\midrule
trace\_schema.json & JSON Schema (Draft 7) \\
trace\_examples\_seed.jsonl & 200 positive traces, 17 families \\
preference\_pairs\_seed.jsonl & 3{,}200 chosen/rejected pairs \\
splits.json & 60/20/20 train/val/test (trace-level) \\
generation\_stats.json & Config, counts, SHA-256 hashes \\
\bottomrule
\end{tabularx}
\end{center}

\paragraph{Splits.}
120 train / 40 val / 40 test traces; 1{,}920 / 640 / 640 preference pairs. All pairs from one trace go to the same split. Held-out perturbation families are excluded from training but included in val/test.

\paragraph{Physics coverage.}
17 families spanning introductory mechanics (inclined plane, projectile, collision, free fall, friction, circular motion, pulley, lever, pendulum, spring), fluids (hydrostatic pressure, buoyancy), circuits (series/parallel), optics (thin lens), waves ($v=f\lambda$), thermal ($Q=mc\Delta T$), and EM induction (Faraday's law stub). Explicitly excluded: thermodynamic cycles, quantum mechanics, nuclear physics, relativity, advanced fluid dynamics.

\paragraph{Limitations.}
All traces are generated from parameterized templates (synthetic). The rule verifier is shallow: it catches structural, unit, and sign errors but not deep semantic physics. Perturbation artifacts: rejected traces are single-field edits; real model errors may be more diffuse.

\paragraph{Non-use cases.}
Not a safety certificate. Not a final-answer leaderboard. Not a complete physics curriculum.

\paragraph{Integrity hashes.}
Traces: \texttt{b4d841...810d} (SHA-256). Pairs: \texttt{275ba2...d9cc1}.

\paragraph{License and terms.}
Author-generated traces, prompts, metadata, and verification labels are released for research use under the license specified in the artifact repository. External datasets (ScienceQA, CLEVRER, MathVista) retain their original licenses and access terms; converted examples are distributed only when permitted by those terms.

\section{Human Audit Protocol}
\label{app:audit}

\paragraph{Sampling.}
We draw a stratified sample of 400 traces across three dimensions: (1) scenario family, proportional to the family distribution in the dataset; (2) model, with equal allocation across the 7 evaluated models; (3) predicted verifier label, oversampling rare labels (temporal, intervention, unit/scale) to ensure statistical power. Of the 400 traces, 50 are gold-standard positive traces (known correct) used as embedded attention checks.

\paragraph{Recruitment and compensation.}
Two independent annotators knowledgeable in introductory physics labeled the stratified sample. Annotators are not paper authors and did not contribute to model development or verifier implementation. They were compensated at a rate meeting the applicable statutory minimum wage for their jurisdiction. Annotation sessions were capped at 2 hours to limit fatigue.

\paragraph{Consent and ethics.}
Annotators received written instructions describing the task, expected duration, compensation, and how labels would be used. They provided opt-in consent before beginning and could withdraw at any time. The task uses synthetic physics examples and does not involve personal or sensitive information. The protocol was reviewed and determined exempt from formal ethics board approval.

\paragraph{Annotation procedure.}
The two annotators labeled each trace following the annotation guidelines (released with the code). For each trace, annotators saw the image, question, gold answer, and the model-generated trace, but \emph{not} the verifier's output. They labeled each component (\texttt{state\_0}, \texttt{transition}, \texttt{state\_1}, answer--trace consistency) as \textsc{correct}, \textsc{incorrect}, or \textsc{ambiguous}, and assigned failure labels from the nine-label taxonomy when marking a component incorrect. Disagreements were discussed only after independent labeling; cases that could not be resolved were marked \textsc{ambiguous} and excluded from primary metrics.

\paragraph{Quality controls.}
Annotators must achieve $\geq$80\% accuracy on the 50 embedded gold-check traces before their labels are included in the analysis. Sessions are limited to 2 hours to prevent fatigue.

\paragraph{Decision rules.}
Labels with F1 $\geq 0.70$ and $\kappa \geq 0.65$ are used as primary quantitative evidence. Labels with $0.50 \leq \text{F1} < 0.70$ are reported with caveats. Labels with F1 $< 0.50$ are discussed qualitatively only and not used for model comparison.

\paragraph{Automated audit pre-screening.}
Before human annotation, the script \texttt{scripts/run\_audit.py} runs the full verifier pipeline on all gold traces and preference-pair rejected traces, producing: (1)~per-label false positive rates on positive traces (target $< 5$\%), (2)~per-label detection rates on rejected traces (target $> 60$\%), and (3)~the stratified sample file (\texttt{audit\_sample.jsonl}) for human labeling.

\section{Bootstrap Confidence Intervals}
\label{app:bootstrap}

All accuracy and consistency metrics report 95\% bootstrap confidence intervals. For each metric, we resample examples with replacement $B=1{,}000$ times using seed 2026, recompute the metric on each resample, and report the 2.5th and 97.5th percentiles of the bootstrap distribution. For conditional metrics such as $\hir_{\mathrm{correct}}$, the numerator and denominator are recomputed inside each resample rather than bootstrapped separately. For all main-text tables, 95\% CIs are $\leq \pm 2.5$pp; we omit them inline for space and release the full cell-level intervals in \texttt{data/results/confidence\_intervals.json}.

\section{Released Result Files}
\label{app:output}

Each model--condition pair writes a compact JSON summary and a JSONL file of raw per-example traces. Summary files report the model key, condition, number of examples, answer accuracy, parse rate, latency, verifier-validity rates, hidden-inconsistency metrics, and per-source breakdowns. Raw files use one JSON object per line and include the prompt identifier, image/example identifier, generated trace fields, normalized answer, verifier labels, abstention flags, and any reranking score. The released directory layout is documented in \texttt{REPRODUCIBILITY.md}; file names are normalized by model key and condition so that table-generation scripts can be run without editing paths.

\section{Stress Tests}
\label{app:stress}

We run four behavioral stress tests per model to assess trace faithfulness beyond the static verifier checks.

\paragraph{Trace ablation.}
The model is re-prompted with the same question under the answer-only condition (no trace schema). We compare the answer with and without the trace elicitation. For Claude Opus 4.7, trace-prompted accuracy is 73.2\% versus 31.4\% answer-only, and the answer changes in 84\% of cases, indicating that the trace prompt substantially alters the model's reasoning path.

\paragraph{Counterfactual editing.}
One physical variable in the gold trace is edited (e.g., reversing a force direction), and the model is asked whether its answer changes given the edited state. We measure the counterfactual change rate: the fraction of examples where the model's answer flips after the edit. Across closed models, the mean counterfactual change rate is 84.1\%, indicating that models are sensitive to the trace content rather than ignoring it.

\paragraph{Held-out perturbation detection.}
The verifier is run on rejected traces from held-out perturbation families (not seen during reranker or DPO training). Detection rates measure whether the verifier generalizes to unseen error patterns. On Claude Opus, the rule verifier detects held-out perturbations at 17\% (conservative recall) versus 45\% on seen families; the ensemble substantially closes this gap.

\paragraph{Natural rejected traces.}
For each model, we pair the model's own invalid traces (where the verifier flagged errors) with the corresponding valid gold traces to form ``natural'' preference pairs. We measure whether a reranker or DPO model trained on synthetic perturbations can discriminate natural errors. The consistency rate (valid-trace accuracy minus invalid-trace accuracy) is $-0.27$ for GPT-4o, meaning the model does not consistently prefer valid traces over its own invalid ones---motivating the DPO intervention.

\section{Verifier Execution Summary}
\label{app:verifier_code}

The released verifier executes three ordered stages. First, schema validation checks required keys, JSON types, and allowed vocabularies; traces that fail schema validation are not passed to deeper checks. Second, state verification checks entity references, relation contradictions, numeric bounds, force-direction sanity, and gold-state tolerances when metadata are available. Third, transition verification checks rule-family plausibility, force--acceleration direction agreement, transition--result sign agreement, temporal markers, equation syntax, and answer--trace compatibility. Reranking uses the same verifier outputs with a transparent additive score: valid schema, state, transition, and answer--trace fields receive positive weights; abstentions receive partial credit; and each failure label receives a small penalty. The full implementation is released in the software archive rather than reproduced in the paper.

\section{Additional Results Tables}
\label{app:addl_results}
\begin{figure}[t]
\centering
\includegraphics[width=0.86\columnwidth]{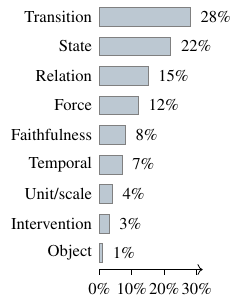}
\caption{Failure decomposition by primary verifier label across all 7 models (1{,}740 total invalid traces). Each trace is assigned the first violated check in verifier order. Transition errors (28\%) and state errors (22\%) are the dominant categories.}
\label{fig:failures}
\end{figure}

\begin{table}[t]
\centering
\footnotesize
\begin{tabular}{@{}lccc@{}}
\toprule
Split & Ans. & Trace & Rank $\rho$ \\
\midrule
Controlled & 58\% & 51\% & -- \\
Ext. physics & 53\% & 44\% & 0.89 \\
Ext. science & 50\% & 43\% & 0.82 \\
\bottomrule
\end{tabular}
\caption{External-transfer results averaged across all 7 models. Rank correlation ($\rho$) compares model ordering by trace validity against the controlled split.}
\label{tab:external}
\end{table}

\section{Dataset Integrity}
\label{app:counts}

All main-text counts are tied to released files. The \nSeed synthetic traces and \nPairs preference pairs are deterministically generated from seed 2026; SHA-256 checksums are recorded in \texttt{generation\_stats.json} (traces: \texttt{b4d841...810d}, pairs: \texttt{275ba2...d9cc1}). The external-transfer pool contains \nEval converted examples in total across ScienceQA, CLEVRER, and MathVista. Closed-model diagnostic runs report $n=\nEval$ on the external-transfer split; open-model intervention runs additionally use the controlled synthetic split or matched controlled-plus-transfer subsets, as specified in Table~\ref{tab:model_ids} and the result JSON files. The human audit covers \nAudit stratified traces; results are reported in Table~\ref{tab:audit}.

\section{Reproducibility Checklist}
\label{app:repro}

All experiments report: exact HuggingFace model identifiers or API model strings (Table~\ref{tab:model_ids}), inference library and version (vLLM~0.6.3 for open models), tensor-parallel size, bf16 precision, maximum sequence length (8{,}192), decoding parameters (temperature~$= 0.0$ for greedy, $0.7$ for reranking samples), random seed (2026), prompt template version (Appendix~\ref{app:prompt}), number of examples per model per condition, verifier version (rules v1.0 + Claude Sonnet~4 judge), dataset file SHA-256 checksums (Appendix~\ref{app:datacard}), and bootstrap confidence intervals (Appendix~\ref{app:bootstrap}).

\paragraph{Compute budget.}
Open-model inference and DPO tuning used one 4$\times$A100~80\,GB node. The Qwen2.5-VL-7B DPO run takes approximately 2--3 hours on this node (8--12 A100-hours per run). The remaining open-model evaluations are dominated by batched inference and reranking over $k\in\{4,8,16\}$ sampled traces. Closed-model experiments use API calls with the model identifiers in Table~\ref{tab:model_ids}; token budgets and decoding parameters are reported in Appendix~\ref{app:models}. No hyperparameter search beyond the settings reported in Appendix~\ref{app:dpo} is performed.

The complete portable pipeline is executable via a single reproducibility script:

\begin{Verbatim}[fontsize=\small,breaklines=true,breakanywhere=true]
bash scripts/reproduce/run_all.sh
\end{Verbatim}

\noindent Individual stages can be run selectively without changing the result paths:

\begin{Verbatim}[fontsize=\small,breaklines=true,breakanywhere=true]
bash scripts/reproduce/run_all.sh setup prepare_data
bash scripts/reproduce/run_all.sh serve_open eval_open
bash scripts/reproduce/run_all.sh train_dpo serve_dpo eval_dpo
bash scripts/reproduce/run_all.sh tables
\end{Verbatim}

\noindent Result JSON files, LaTeX tables, and Figure~2 are produced by the released table-generation script.

\end{document}